\documentclass[11pt,a4paper]{article}
\usepackage[hyperref]{ranlp2021}
\usepackage{times}
\usepackage{latexsym}

\usepackage{microtype}

\usepackage{kotex}
\usepackage{kotex-logo}
\usepackage{multirow}
\usepackage{graphicx}
\usepackage{bm}
\usepackage{amsmath}

\title{Transforming Multi-Conditioned Generation from Meaning Representation}

\date{}

\begin{document}
\maketitle
\begin{abstract}
In task-oriented conversation systems, natural language generation systems that generate sentences with specific information related to conversation are useful. Our study focuses on language generation by considering various information representing the meaning of utterances as multiple conditions of generation. Data-to-text generation, the conditions for sentence meaning, generally goes through two steps: sentence planning and surface realization. However, we propose a simple one-stage framework to generate utterances directly from MR (Meaning Representation). Our model is based on GPT2 and generates utterances with flat conditions on slot and value pairs, which does not need to determine the structure of the sentence. We evaluate several systems in the E2E dataset with 6 automatic metrics. Our system is a simple method, but it demonstrates comparable performance to previous systems in automated metrics. In addition, using only 10\% of the dataset without any other techniques, our model achieves comparable performance, and shows the possibility of performing zero-shot generation and expanding to other datasets.
\end{abstract}

\section{Introduction}
In many conversation systems, generating sentences with specific information is useful. For example, it can be used in chatbot systems or spoken dialogue systems to generate utterances that contain \textit{meaning representations} (MRs) corresponding to a user's query. In order to train the NLG system that reflects this variety of information, a large amount of labeled data is required. At the 2017 E2E challenge~\citep{dusek2019e2e}, a large dataset was released, which consisted of pairs of MRs representing restaurant reviews and corresponding utterances. Table~\ref{Tab:sample} shows an example. MRs can be regarded as the multi-conditions type of utterance generation, which consists of slots and values, and the corresponding utterances are references written by humans. We focus on training the model to generate utterances directly from MRs. 

\begin{table}[]
\centering
\resizebox{1.0\columnwidth}{!}{
\begin{tabular}{|c|c|}
\hline
{\color[HTML]{000000} MR   (slot{[}value{]})} & {\color[HTML]{000000} \begin{tabular}[c]{@{}c@{}}name{[}Giraffe{]}, eatType{[}pub{]}, food{[}Fast food{]},\\       area{[}riverside{]}, familyFriendly{[}yes{]}\end{tabular}} \\ \hline
{\color[HTML]{000000} Utterance}              & {\color[HTML]{000000} \begin{tabular}[c]{@{}c@{}}On the riverside the Giraffe is \\ a Fast food, kid friendly pub.\end{tabular}}                                              \\ \hline
\end{tabular}}
\caption{An example of an E2E dataset consists of pairs of a MR and an utterance. The purpose of the E2E challenge is to create a system that generates utterance that reflects MR.}
\label{Tab:sample}
\end{table}

Many of the previous NLG research are a two-stage approach through sentence planning and surface realization. Sentence planning determines the overall sentence structure and surface realization is the process of flattening the sentence structure. In recent studies~\citep{konstas2013global, dusek_jurcicek_2016_sequence, juraska_etal_2018_deep}, the neural network model processes these two-stage at once by training end-to-end without the need for aligned data.

When generating sentences from an input (flat or structured MR), there are a template-based approach and a neural network-based approach. \citet{smiley_etal_2018_e2e, puzikov_gurevych_2018_e2e, wiseman_etal_2018_learning} generate sentences based on template. The template method is to obtain the structural sets of sentences corresponding to MRs from training data and apply the template appropriate to the test data to generate the sentences. In \citet{smiley_etal_2018_e2e, puzikov_gurevych_2018_e2e}, a template is formed based on rules, and \citet{wiseman_etal_2018_learning} learns the template structure of a sentence as a neural network.

There is also a way to generate a natural language with a neural network without using a template. \citet{dusek_jurcicek_2016_sequence, smiley_etal_2018_e2e, puzikov_gurevych_2018_e2e, juraska_etal_2018_deep, elder_etal_2019_designing, gehrmann_etal_2018_end} are sequence-to-sequence models of encoders and decoders, which have a one-stage framework, and ~\citet{dusek_jurcicek_2016_sequence} is the model used by E2E challenge organizers. In \citet{balakrishnan_etal_2019_constrained}, the encoder and decoder model converts flat MRs to structure MRs, and outputs the sentence through constrained decoding.

Template and neural network methods each have advantages and disadvantages. The template method guarantees a certain performance but limits the diversity and possibilities of the output. The neural network method generally performs better than the template method, but requires a lot of data and has limitations in the naturalness and semantical correctness of sentences~\citep{nayak2017plan}.

We propose a novel approach using the Transformer decoder as a simple one-stage framework. In our model, GPT2-small~\citep{radford2019language} is the backbone, and $(s_i, v_i)$ pairs of the meaning representation are put into multiple conditions to generate a sentence. In the previous works~\citep{juraska_etal_2018_deep, balakrishnan_etal_2019_constrained, smiley_etal_2018_e2e, puzikov_gurevych_2018_e2e, dusek_jurcicek_2016_sequence}, when receiving the meaning representation as input, the value is delexicalized. Specifically, all values corresponding to the same slot are delexicalized to a placeholder so that unseen inputs can be processed. However, \citet{nayak2017plan, juraska_etal_2018_deep} report that delexicalization often leads to inappropriate behavior in scenarios. For example, the word "cheap" can be reflected in the utterance that matches the value of "less than \$20". Also, when food[Italian] is given as slot[value], "Italian food" is an appropriate phrase in a generated utterance, but in the case of food[fast food], "fast food food" is an incorrect phrase. Additionally, the combination of eatType[coffee shop] and food[Italian] is rather weird, and the combination of name[The Rice Boat] and area[riverside] is appropriate, so delexicalization doesn't fully utilize the characteristics of tokens. We, therefore, treat the slot as a special token and the value as a regular token of vocabulary without delexicalization in the training. Nevertheless, in testing for unseen values, the model generates appropriate utterances and is described in Section~\ref{sec:zero-shot}. Our method is considered as generating a sentence as a simple one-stage framework directly from flat MRs.

Our model is tested on the E2E dataset. In addition to the evaluation metrics used in the E2E challenge, the systems are evaluated with BERTScore~\citep{Zhang2020BERTScore}. BERTScore is an evaluation metric used in general text generation. It obtains a semantic similarity score between two sentences. Our approach shows the best performance in BLEU, METEOR, and BERTScore and competitive performance in other metrics. By leveraging the pre-trained model GPT2, we quickly converge the model with only a few epochs and generate fluent utterances without considering the structure of the sentence. In addition, even if only 10\% of the training data is used, it achieves performance comparable to previous systems.

\section{Related Work}
In many NLP tasks, the Transformer-based~\citep{vaswani2017attention} models have recently shown good performance. ~\citet{gehrmann_etal_2018_end} is a previous work using the Transformer encoder and decoder in the E2E task. BERT~\citep{bert} composed only of the Transformer encoder is used a lot in NLU tasks, and GPT~\citep{radford2019language} composed only of the Transformer decoder is used a lot in NLG tasks. These models are trained as large-scale open-domain corpora. By leveraging pre-trained models trained on large datasets and applying them to downstream tasks, many NLP tasks achieve better performance.

The generation of utterances from MRs is quite similar to machine translation, one of the \textit{sequence-to-sequence} tasks. Also, in terms of generating sentences with certain restrictions, it is similar to style transfer~\citep{logeswaran2018content, lample2018multipleattribute, lee_2020_stable}, which is one of \textit{sequence+condition-to-sequence} tasks. However, the generation of utterances from MRs is not exactly the same as the above tasks because of the \textit{condition-to-sequence} perspective. Since the E2E task does not have a given sequence as the input of the model, we approach the sentence generation using only the decoder without sequence encoding. We choose a language model of the Transformer decoder that performs better than LSTM and uses the pre-trained model GPT2 as a backbone.

\section{Our Approach}

\subsection{Problem Statement}
\paragraph{E2E dataset} The domain of the E2E dataset is a restaurant and consists of $D=\{(\bm{M}_1, u_1), \cdots, (\bm{M}_n, u_n) \}$. $\bm{M}_i$ is the MR and contains the (slot, value) pair, $(s_i, v_i)$, and $ u_i $ is the corresponding utterance. There are 8 types of slots, and the value corresponding to each slot has various numbers (2 to 34). Statistics of the overall dataset are shown in Table~\ref{Tab:statistic}.

The goal of the task is to generate a $u_i$ by reflecting $\bm{M}_i$. $s_i$ is the concept of a given category of conditions, and $v_i$ is an item that should actually be reflected in utterance generation. Table~\ref{Tab:failed slot} shows that the utterance may not include $v_i$, which may have an incomplete reason for the data, but the main reason is also the reason that it was reflected as a synonym for $v_i$. Table~\ref{Tab:insufficient} shows that the percentage of values that are not equally in utterance are 15.2\% and 11.58\% in training and test data, respectively (except for \textit{familyfriendly} because of boolean type). Therefore, changing the value to a placeholder assumes that there is a risk that the input and output will not match properly. In model training, not only delexcalization but also preprocessing of special data such as \citet{smiley_etal_2018_e2e} is not performed, and the model is expected to learn about tokens and phrases with different inputs and outputs such as synonyms.

\begin{table}[]
\centering
\resizebox{1.0\columnwidth}{!}{
\begin{tabular}{|c|c|c|c|c|}
\hline
{\color[HTML]{000000} E2E Dataset} & {\color[HTML]{000000} MRs}   & {\color[HTML]{000000} References} & {\color[HTML]{000000} Slots/MR} & {\color[HTML]{000000} Tokens/Ref} \\ \hline
{\color[HTML]{000000} training}    & {\color[HTML]{000000} 4,862} & {\color[HTML]{000000} 42,061}     & {\color[HTML]{000000} 5.52}     & {\color[HTML]{000000} 20.27}      \\ \hline
{\color[HTML]{000000} develpoment} & {\color[HTML]{000000} 547}   & {\color[HTML]{000000} 4,672}      & {\color[HTML]{000000} 6.3}      & {\color[HTML]{000000} 24.52}      \\ \hline
{\color[HTML]{000000} test}        & {\color[HTML]{000000} 630}   & {\color[HTML]{000000} 4,693}      & {\color[HTML]{000000} 6.91}     & {\color[HTML]{000000} 26.76}      \\ \hline
\end{tabular}
}
\caption{Statistics of E2E dataset provided by \citet{duvsek2020evaluating}.}
\label{Tab:statistic}
\end{table}

\begin{table}[]
\centering
\resizebox{1.0\columnwidth}{!}{
\begin{tabular}{|c|c|}
\hline
{\color[HTML]{000000} MR}       & {\color[HTML]{000000} \begin{tabular}[c]{@{}c@{}}name{[}The Rice   Boat{]}, food{[}English{]},\\       \color[HTML]{3166FF} priceRange{[}less   than £20{]}, \color[HTML]{FE0000} customer   rating{[}low{]},\\      area{[}riverside{]}, familyFriendly{[}yes{]}, near{[}Express by Holiday Inn{]}\end{tabular}} \\ \hline
{\color[HTML]{000000} Utterance} & {\color[HTML]{000000} \begin{tabular}[c]{@{}c@{}}The   family friendly The Rice Boat is located on the riverside,\\      near the Express by Holiday Inn,\\       serving English cuisine \color[HTML]{3166FF}  below £20 \color[HTML]{000000} with a \color[HTML]{FE0000} customer rating.\end{tabular}}                               \\ \hline
\end{tabular}
}
\caption{Blue indicates that the value is reflected as an expression of a synonym, and red indicates that the value is not reflected due to data error.}
\label{Tab:failed slot}
\end{table}

\begin{table}[t]
\centering
\resizebox{0.6\columnwidth}{!}{
\begin{tabular}{|c|c|}
\hline
{\color[HTML]{000000} Type}                                          & {\color[HTML]{000000} not included(\%)} \\ \hline
{\color[HTML]{000000} training dataset}                              & {\color[HTML]{000000} 15.22}          \\ \hline
{\color[HTML]{000000} test dataset}                                  & {\color[HTML]{000000} 11.58}          \\ \hline
\end{tabular}
}
\caption{Ratio where the value of MR does not exist as it is in the utterance}
\label{Tab:insufficient}
\end{table}

\begin{figure*}[!t]
    \centering 
    \includegraphics[width=2.0\columnwidth]{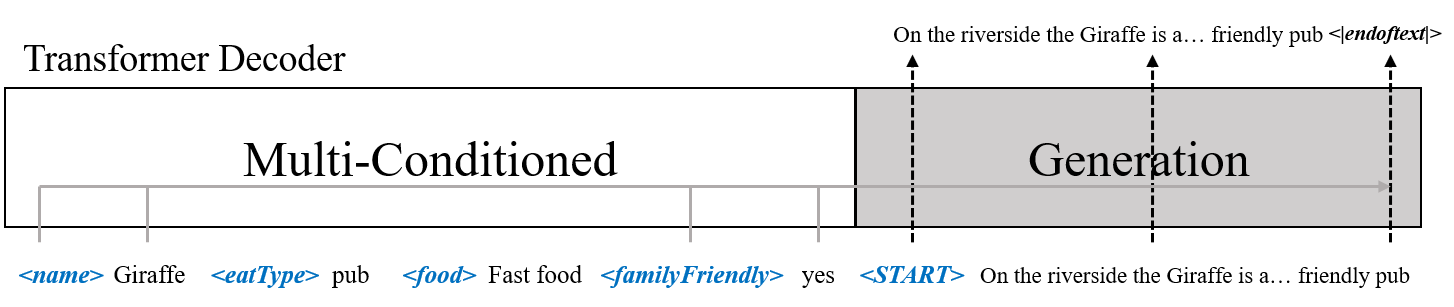}
    \caption{Our structure has GPT2 backbone based on the Transformer decoder. Blue tokens are special tokens that are newly added to the vocabulary. The model autoregressively starts to generate utterance from when the \textit{\textless{}START\textgreater{}} token is received as input, and ends when the $\textless{}\left | endoftext \right |\textgreater{}$ token is output.}
    \label{fig:model}
\end{figure*}

\subsection{Pre-trained Model: GPT2-small}
Our system uses GPT2-small model as a backbone to generate utterances and is illustrated in Figure~\ref{fig:model}. GPT2 is an unsupervised pre-trained model with large-scale open-domain corpora of unlabeled text. GPT2 uses only the Transformer decoder and generates sentences from left to right autoregressively without an encoder. GPT2-small has 768-dimensional embedding size, 12 heads, and 12 layers, so the total number of parameters is 117M. Our system has the advantage of starting from knowing the token distribution by using the pre-trained model as the initial state.

\subsection{Generation}
As conditions of an utterance generation, $(s_i, v_i)$ pairs of a flat MR are given, slots are treated as special tokens. Specifically, the tokens for $\textit{slots} $ and $\textit{start} $ are added to vocabulary, and the regular tokens in vocabulary are used for $\textit{values}$ and $\textit{end}$. In order to distinguish between special tokens and regular tokens, special tokens are changed to \textit{\textless{}SPECIAL TOKEN\textgreater{}}\ to form vocabulary. Inputs are given by concatenating the given $(s_i, v_i)$ pairs in order (e.g. \textit{\textless{}name\textgreater{}}, Giraffe, \textit{\textless{}eatType\textgreater{}}, pub, ...). We use \textit{\textless{}name\textgreater{}}, \textit{\textless{}eatType\textgreater{}}, \textit{\textless{}food\textgreater{}}, \textit{\textless{}priceRange\textgreater{}}, \textit{\textless{}customer rating\textgreater{}}, \textit{\textless{}area\textgreater{}}, \textit{\textless{}familyFriendly\textgreater{}}, \textit{\textless{}near\textgreater{}} in the fixed order of slots as formed in the E2E dataset, and do not put slots not in the given MR as input. Our system generates utterances by considering the MR as multi-conditions of generation. Through training end-to-end Multi-Conditioned generation, we hope that the model find out the role of a MR.

Transformer Decoder is autoregressively unidirectional from left to right, so the model only sees the previous tokens:
\begin{equation}
\label{TransformerDecoder}
\bm{o_{i}} = \textrm{TransformerDecoder} (u_{i}^{1:k}, \textit{\scriptsize{\textless{}START\textgreater{}}}, s_i, v_i)
\end{equation}
where $u_{i}^{1:k}$ denotes tokens up to $k^{th}$ tokens of $u_i$. $\bm{o_{i}}$ is the output of Transformer Decoder, $k \times h$ tensor, and $h$ is the hidden embedding dimension of the decoder.

To predict the token of the next step $(k+1)$, multiply the $k^{th}$-vector $o_{i}^{k}$ by matrix $M$ as follows:
\begin{equation}
\label{output}
u_{i}^{k+1} = argmax(M(o_{i}^{k}))
\end{equation}
where $M$ is a randomly initialized matrix.

Because our system is a one-stage framework that generates utterance directly from flat MRs, sentence planning and surface realization are not considered separately. Our approach shows strong results in~\ref{sec:results} without additional techniques such as delexicalization, data augmentation, and extra datasets. When testing, it is possible to deal with unseen values by delexicalization, which is described in detail in Section~\ref{sec:zero-shot}. In addition, Section~\ref{sec:small-MRs} introduces utterance generation for $(s_i, v_i)$ pairs that exist in the real world even if they are not in the training dataset.

\subsection{Training}
We experiment using one V100 16GB GPU in Linux environment on an AWS server. Our system is end-to-end trained with the AdamW~\citep{loshchilov2018decoupled} optimizer for 5 epochs. The initial value of the learning rate is 2e-5 and is adjusted with a linear scheduler. The model is trained so that the output at the current step $(k)$ predicts the token of the next step $(k+1)$ and the loss of the objective function is calculated as:
\begin{equation}
\begin{split}
\label{loss}
\mathcal{L}(\theta) &= -\sum_{(\bm{M}_i, v_i) \in D} (\mathrm{log}\,p(u^{1}|\textit{\scriptsize{\textless{}S\textgreater{}}}, s_i, v_i) \\
&+ \sum_{k=1} \mathrm{log}\,p(u_{i}^{k+1} | u_{i}^{1:k}, \textit{\scriptsize{\textless{}S\textgreater{}}}, s_i, v_i) \\
&+ \mathrm{log}\,p(\textit{\scriptsize{\textless{} E \textgreater{}}} |u_{i}, \textit{\scriptsize{\textless{}S\textgreater{}}}, s_i, v_i))
\end{split}
\end{equation}
where $u_{i}^{k}$ is the $k^{th}$ tokens of $u_i$ and \textit{\small{\textless{}S\textgreater{}}}, \textit{\small{\textless{}E\textgreater{}}} are \textit{start}, \textit{end} token respectively. Since the ground truth given in the dataset is only utterance, the outputs before \textit{\small{\textless{}S\textgreater{}}} is entered cannot be used for loss calculation.

During training time, tokens are generated by applying teacher-forcing, and tokens are generated by self-feeding during testing time.

\section{Experiments}
We use the model provided by HuggingFace~\footnote{https://huggingface.co/gpt2} to make it easy to use the pre-trained GPT2 trained by OpenAI.

\subsection{Evaluation Metrics}
We use the five automatic evaluation metrics used in the E2E Challenge, BLEU~\citep{papineni2002bleu}, NIST~\citep{lin2004automatic}, METEOR~\citep{denkowski2014meteor}, ROUGE~\citep{rouge} and CIDEr~\citep{vedantam2015cider}, equally as the basis. Evaluation scripts are provided by challenge organizers~\footnote{https://github.com/tuetschek/e2e-metrics}. We additionally calculate the similarity F1 scores of the two sentences using the BERTscore of RoBERTa model~\citep{liu2020roberta} provided by the library~\footnote{https://pypi.org/project/bert-score/}. The two sentences entered in the BERTscore library are the generated utterances and human references. The BERTScore metric is task agnostic and, unlike previous metrics, uses importance weighting between contextual embedding. Therefore, it is a common metric that calculates a better correlation by solving the disadvantages of the previous metric. BERTScore is measured only for the systems that provided the output for the test dataset.

\subsection{Results}
\label{sec:results}

\begin{table*}[t]
\centering
\resizebox{1.75\columnwidth}{!}{
\begin{tabular}{|c|cccccc|} 
\hline
{\color[HTML]{000000} System}                                   & {\color[HTML]{000000} BLEU}                                                      & {\color[HTML]{000000} NIST}                                                      & {\color[HTML]{000000} METEOR}                                                    & {\color[HTML]{000000} ROUGE\_L}                                                  & {\color[HTML]{000000} CIDEr}                                                     & {\color[HTML]{000000} BERTscore}                                                 \\ \hline\hline
{\color[HTML]{000000} Multi-Conditioned   Transformer} & {\color[HTML]{000000} \textbf{0.6794}}                                           & {\color[HTML]{000000} 8.6477}                                                    & {\color[HTML]{000000} \textbf{0.4579}}                                           & {\color[HTML]{000000} 0.6998}                                                    & {\color[HTML]{000000} 2.2884}                                                    & {\color[HTML]{000000} \textbf{0.942}}                                            \\ \hline
{\color[HTML]{000000} 30\% sampling   (avg)}                    & {\color[HTML]{000000} \begin{tabular}[c]{@{}c@{}}0.6651\\ \footnotesize{(0.002)}\end{tabular}}  & {\color[HTML]{000000} \begin{tabular}[c]{@{}c@{}}8.5712\\ \footnotesize(0.035)\end{tabular}} & {\color[HTML]{000000} \begin{tabular}[c]{@{}c@{}}0.4364\\ \footnotesize(0.0056)\end{tabular}} & {\color[HTML]{000000} \begin{tabular}[c]{@{}c@{}}0.6871\\ \footnotesize(0.0022)\end{tabular}} & {\color[HTML]{000000} \begin{tabular}[c]{@{}c@{}}2.1561\\ \footnotesize(0.0360)\end{tabular}} & {\color[HTML]{000000} \begin{tabular}[c]{@{}c@{}}0.940\\ \footnotesize(0.00067)\end{tabular}} \\ \hline
{\color[HTML]{000000} 10\% sampling   (avg)}                    & {\color[HTML]{000000} \begin{tabular}[c]{@{}c@{}}0.6541\\ \footnotesize(0.0024)\end{tabular}} & {\color[HTML]{000000} \begin{tabular}[c]{@{}c@{}}8.4332\\ \footnotesize(0.043)\end{tabular}}  & {\color[HTML]{000000} \begin{tabular}[c]{@{}c@{}}0.4271\\ \footnotesize(0.0033)\end{tabular}} & {\color[HTML]{000000} \begin{tabular}[c]{@{}c@{}}0.6761\\ \footnotesize(0.0039)\end{tabular}} & {\color[HTML]{000000} \begin{tabular}[c]{@{}c@{}}2.0786\\ \footnotesize(0.0354)\end{tabular}} & {\color[HTML]{000000} \begin{tabular}[c]{@{}c@{}}0.939\\ \footnotesize(0.00045)\end{tabular}} \\ \hline
{\color[HTML]{000000} No pre-trained}                          & {\color[HTML]{000000} 0.5885}                                                    & {\color[HTML]{000000} 8.0320}                                                    & {\color[HTML]{000000} 0.3962}                                                    & {\color[HTML]{000000} 0.6302}                                                     & {\color[HTML]{000000} 1.7585}                                                    & {\color[HTML]{000000} 0.930}                                                     \\ \hline\hline
{\color[HTML]{000000} Tgen (baseline)}                          & {\color[HTML]{000000} 0.6593}                                                    & {\color[HTML]{000000} 8.6094}                                                    & {\color[HTML]{000000} 0.4483}                                                    & {\color[HTML]{000000} 0.685}                                                     & {\color[HTML]{000000} 2.2338}                                                    & {\color[HTML]{000000} 0.939}                                                     \\ \hline
    {\color[HTML]{000000} Model-T}                                  & {\color[HTML]{000000} 0.5657}                                                    & {\color[HTML]{000000} 7.4544}                                                    & {\color[HTML]{000000} 0.4529}                                                    & {\color[HTML]{000000} 0.6614}                                                    & {\color[HTML]{000000} 1.8206}                                                    & {\color[HTML]{000000} 0.938}                                                     \\ \hline
{\color[HTML]{000000} Slug2Slug}                                & {\color[HTML]{000000} 0.6619}                                                    & {\color[HTML]{000000} 8.613}                                                     & {\color[HTML]{000000} 0.4454}                                                    & {\color[HTML]{000000} 0.6772}                                                    & {\color[HTML]{000000} 2.2615}                                                    & {\color[HTML]{000000} \textbf{0.942}}                                            \\ \hline
{\color[HTML]{000000} TemplGen}                                 & {\color[HTML]{000000} 0.4202}                                                    & {\color[HTML]{000000} 6.7686}                                                    & {\color[HTML]{000000} 0.3968}                                                    & {\color[HTML]{000000} 0.5481}                                                    & {\color[HTML]{000000} 1.4389}                                                    & {\color[HTML]{000000} -}                                                         \\ \hline
{\color[HTML]{000000} SeqGen}                                   & {\color[HTML]{000000} 0.6336}                                                    & {\color[HTML]{000000} 8.1848}                                                    & {\color[HTML]{000000} 0.4322}                                                    & {\color[HTML]{000000} 0.6828}                                                    & {\color[HTML]{000000} 2.1425}                                                    & {\color[HTML]{000000} -}                                                         \\ \hline
{\color[HTML]{000000} NTemp+AR}                                 & {\color[HTML]{000000} 0.598}                                                     & {\color[HTML]{000000} 7.56}                                                      & {\color[HTML]{000000} 0.3875}                                                    & {\color[HTML]{000000} 0.6501}                                                    & {\color[HTML]{000000} 1.95}                                                      & {\color[HTML]{000000} -}                                                         \\ \hline
{\color[HTML]{000000} dot, copy, K =   2}                       & {\color[HTML]{000000} 0.674}                                                     & {\color[HTML]{000000} 8.61}                                                      & {\color[HTML]{000000} 0.452}                                                     & {\color[HTML]{000000} 0.708}                                                     & {\color[HTML]{000000} 2.31}                                                      & {\color[HTML]{000000} -}                                                         \\ \hline
{\color[HTML]{000000} Transformer, K   = 2}                     & {\color[HTML]{000000} 0.662}                                                     & {\color[HTML]{000000} 8.6}                                                       & {\color[HTML]{000000} 0.457}                                                     & {\color[HTML]{000000} 0.704}                                                     & {\color[HTML]{000000} \textbf{2.34}}                                                      & {\color[HTML]{000000} -}                                                         \\ \hline
{\color[HTML]{000000} TripAdvisor}                              & {\color[HTML]{000000} 0.6738}                                                    & {\color[HTML]{000000} \textbf{8.7277}}                                           & {\color[HTML]{000000} 0.4572}                                                    & {\color[HTML]{000000} \textbf{0.7152}}                                           & {\color[HTML]{000000} 2.2995}                                           & {\color[HTML]{000000} -}                                                         \\ \hline
\end{tabular}
}
\caption{Automatic metric scores of our and compared systems in the E2E test dataset. Systems are evaluated with BERTscore along with the five metrics used in the E2E challenge. Our system, which is trained by sampling the training dataset, is the average of the results of three attempts, and the number in parentheses is the standard deviation. The bold number is a notation for the best performing system.}
\label{Tab:Results}
\end{table*}
Table~\ref{Tab:Results} shows the experimental results of our system and comparison systems. The first section is our model \textit{Multi-Conditioned Transformer}, which consists of the Transformer decoder.

\begin{table*}[t]
\centering
\resizebox{2.0\columnwidth}{!}{
\begin{tabular}{|c|c|}
\hline
{\color[HTML]{000000} MR}                                       & {\color[HTML]{000000} name{[}Blue   Spice{]}, eatType{[}pub{]}, area{[}riverside{]}}                                                                                                                                                                                        \\ \hline
{\color[HTML]{000000} \textbf{Multi-Conditioned   Transformer}}  & {\color[HTML]{000000} \textbf{Blue Spice is a pub   in the riverside area.}}                                                                                                                                                                                                \\ \hline
{\color[HTML]{000000} Tgen   (baseline)} & {\color[HTML]{000000} Blue Spice is a pub   by the riverside.}                                                                                                                                                                                                              \\ \hline
{\color[HTML]{000000} Model-T}           & {\color[HTML]{000000} Blue Spice is a pub   located in the riverside area.}                                                                                                                                                                                                 \\ \hline
{\color[HTML]{000000} Slug2Slug}         & {\color[HTML]{000000} Blue Spice is a pub   in the riverside area.}                                                                                                                                                                                                         \\ \hline
{\color[HTML]{000000} Reference sample}                                & {\color[HTML]{000000} There is a pub Blue   Spice in the riverside area.} \\ \hline\hline

{\color[HTML]{000000} MR}                                       & {\color[HTML]{000000} \begin{tabular}[c]{@{}c@{}}name{[}The   Mill{]}, eatType{[}restaurant{]}, food{[}English{]}, priceRange{[}less than £20{]},\\      area{[}city centre{]}, familyFriendly{[}yes{]}, near{[}Raja Indian Cuisine{]}\end{tabular}}                        \\ \hline
{\color[HTML]{000000} \textbf{Multi-Conditioned   Transformer}}  & {\color[HTML]{000000} \textbf{\begin{tabular}[c]{@{}c@{}}The   Mill is a family-friendly restaurant that serves English food for less than   £20.\\      It is located in the city centre near Raja Indian Cuisine.\end{tabular}}}                                          \\ \hline
{\color[HTML]{000000} Tgen   (baseline)} & {\color[HTML]{000000} \begin{tabular}[c]{@{}c@{}}The   Mill is a family-friendly english restaurant in the city centre near \\      Raja Indian Cuisine with a price range less than £20.\end{tabular}}                                                                     \\ \hline
{\color[HTML]{000000} Model-T}           & {\color[HTML]{000000} \begin{tabular}[c]{@{}c@{}}The   Mill is a family-friendly restaurant which serves English food in the price   range of less than £20. \\      It is located in the city centre area, near Raja Indian Cuisine.\end{tabular}}                         \\ \hline
{\color[HTML]{000000} Slug2Slug}         & {\color[HTML]{000000} \begin{tabular}[c]{@{}c@{}}The   Mill is a family friendly English restaurant in the city centre near Raja   Indian Cuisine. \\      It has a price range of less than £20.\end{tabular}}                                                             \\ \hline
{\color[HTML]{000000} Reference sample}                                & {\color[HTML]{000000} \begin{tabular}[c]{@{}c@{}}The   Mill, Is a restaurant and is family-friendly, cheap and reasonable priced is   very good for the family , \\      We provide full English food. Located near Raja Indian Cuisine In the city   centre.\end{tabular}} \\ \hline
\end{tabular}
}
\caption{Comparison of utterances for given MRs. Samples given according to when 3 and 7 $(s_i, v_i)$ pairs are given. In the E2E dataset, the human reference provides several versions but extracts one sample.}
\label{Tab:Generated Samples}
\end{table*}

\subsubsection{Compared Systems}
TGen~\citep{dusek_jurcicek_2016_sequence} is the baseline tested by the E2E challenge organizer. SeqGen~\citep{smiley_etal_2018_e2e} is the system that participated in the challenge, and Slug2Slug~\citep{juraska_etal_2018_deep} is the system that won the E2E challenge. Slug2Slug improves performance by learning the surface realization model as additional data and ensemble the three models. Model-T~\citep{puzikov_gurevych_2018_e2e} and TempleGen~\citep{smiley_etal_2018_e2e} are rule-based systems using templates. NTemp + AR (autoregressive)~\citep{wiseman_etal_2018_learning} is a hidden semi-markov model (HSMM) decoder that learns the structure of a template. Template-based systems guarantee a certain quality and fluency of natural language generation, but overall performance is lower than neural networks. Dot-copy and Transformer~\citep{gehrmann_etal_2018_end} are methods of learning the structure of a template with a neural encoder and decoder. The hyperparameter $K$ of these two systems indicates the number of models to be diverse ensembling. TripAdvisor~\citep{elder_etal_2019_designing} follows a two-stage approach: (1) content selection at the system input to generate a symbol intermediate representation and (2) generating utterance. Each stage proceeds with the structure of a neural encoder and decoder and improves the performance of the model with additional data.

\subsubsection{Automatic Evaluation}
The performance of our system and the comparison systems are shown in Table~\ref{Tab:Results}. In these systems, Multi-Conditioned Transformer achieves the best performance in BLEU, METEOR, and BERTScore, and the second-best in NIST. Our system also shows competitive performance compared to previous systems in ROUGE and CIDEr metrics. We experimented with at least 3 random seeds and observe that our model always reaches similar performance.

"No pre-trained" is a model trained from scratch and the rest are the same except for initialization. If our model is trained without the pre-trained technique, we observe that the model performance is quite degraded. However, our approach simply and effectively derives the generalization performance of the model by using only pre-trained LM without using other techniques such as additional data and ensemble techniques.


\subsubsection{Human Evaluation}
\begin{table}[]
\centering
\resizebox{0.6\columnwidth}{!}{
\begin{tabular}{|c|c|c|}
\hline
model & quality        & naturalness    \\ \hline
ours  & \textbf{4.525} & \textbf{4.625} \\ \hline
TGen  & 4.317          & 4.498          \\ \hline
Slug2Slug  & 4.340          & 4.545          \\ \hline
\end{tabular}
}
\caption{Average of three workers' ratings}
\label{Tab:human-evaluation}
\end{table}
Human evaluation is performed for quality and naturalness as in ~\citet{duvsek2020evaluating, juraska_etal_2018_deep}, and the results are shown in Table~\ref{Tab:human-evaluation}. Quality is a score for grammatical correctness and whether generated utterance properly reflects given MRs. Naturalness is a rating of the possibility that utterance is written by a native speaker, regardless of the MRs. We randomly sampled 200 samples from our test set and hired 3 workers from Amazon Mechanical Turk~\footnote{https://www.mturk.com/} to rate them on a scale of 1(bad)-5(good). Slug2Slug is a system that ranked first and second in quality and naturalness, respectively, in the E2E challenge. In human evaluation, our model shows better than baseline TGen and Slug2Slug.

\subsubsection{Utterance Generation}
Table~\ref{Tab:Generated Samples} shows the output of the systems. As with BERTscore calculations, other comparative systems with no output provided cannot verify the utterances. When there are three $(s_i, v_i)$ pairs, there is no difference between our system and previous systems. However, as the number of pairs increases, the possible sentence structures vary, so different systems output different utterances. Test data with many pairs is considered to make a difference in automatic evaluation performance. Since our system is based on the Transformer, it can be more robust and general to long-term sequences than LSTM-based systems.

\subsubsection{Training with less data}
The 2nd and 3rd rows of Table~\ref{Tab:Results} are the results of fine-tuning our model by sampling only a small amount of training dataset. For small training data, 10\% and 30\% of the entire training dataset are randomly sampled. Our model is trained and averaged by performing random sampling three times in consideration of the possibility that the performance of the model may vary according to the statistics of the randomly sampled dataset. We found that the performance of our model was similar even with random sampling. Calculating the standard deviation for BLEU, NIST, METEOR, ROUGE, CIDEr, and BERTscore, respectively, results in 0.002, 0.0351, 0.0056, 0.0022, 0.0359 0.00067 when 30\% of the training data is used, and 0.0024, 0.043, 0.0033, 0.0039, 0.0354, 0.00045 when 10\% of the training data is used. Our system shows a similar level of performance with a small standard deviation according to the sampled data.

As the sampling of the training data of the system increases, the performance of the system improves. However, if we use more than 50\% of the training data through many experiments, our system has little improvement in performance. Our approach can leverage the pre-trained language model to take advantage of the background knowledge of sentence generation. Therefore, it is difficult to expect a linear relationship between increasing the number of training data and increasing performance. However, with the effect of background knowledge, our system, which was trained by sampling 10\% of the training data, shows performance comparable to previous systems. The performance of the model trained by sampling 30\% of the data is similar to that of the Slug2Slug, the system that won the E2E challenge. Building our system with only 30\% of the training data and showing good results demonstrates the effectiveness of using the pre-trained model. If a better pre-trained model is used as the backbone, we hope to build an effective model with less data.

\subsection{Generation from Unseen Values}
\label{sec:zero-shot}

\begin{table*}[t]
\centering
\resizebox{2.0\columnwidth}{!}{

\begin{tabular}{|c|c|}
\hline
Slots                 & \textless{}name\textgreater{},   \textless{}food\textgreater{}, \textless{}customer rating\textgreater{}, \textless{}area\textgreater{}, \textless{}near\textgreater{} \\ \hline
Unseen values         & Blue Man, hot food,   2.1 out of 5, countryside, the school                                                                                                            \\ \hline
Delexicalization      & Green Man, Fast food,   3 out of 5, city centre, The Bakers                                                                                                            \\ \hline
Generated   utterance & Green Man is a fast   food restaurant in the city centre near The Bakers. It has a customer rating   of 3 out of 5.                                                    \\ \hline
Relexicalization      & Blue Man is a fast   food restaurant in the countryside near the school. It has a customer rating   of 2.1 out of 5.                                                   \\ \hline\hline
Slots                 & \textless{}name\textgreater{},   \textless{}food\textgreater{}, \textless{}priceRange\textgreater{}, \textless{}familyFriendly\textgreater{}                           \\ \hline
Unseen values         & The Guy, German,   expensive, yes                                                                                                                                      \\ \hline
Delexicalization      & The Wrestlers,   Japanese, high, yes                                                                                                                                   \\ \hline
Generated   utterance & The Wrestlers is a   high priced Japanese restaurant that is children friendly.                                                                                        \\ \hline
Relexicalization      & The Guy is a   expensive priced German restaurant that is children friendly.                                                                                           \\ \hline
\end{tabular}

}
\caption{Zero-shot generation from unseen values. These are two examples, and the given unseen values are delexicalized to similar values in the list. The system generates utterances of delexicalized values with multiple conditions.}
\label{Tab:zero-shot}
\end{table*}

Our model is not trained on unseen values, so it can have weaknesses in real applications. Therefore, we introduce a zero-shot generation method through Sim-Delexicalization. Table~\ref{Tab:zero-shot} shows an example of this experiment. The value of \textit{\textless{}familyFriendly\textgreater{}} is not treated as unseen value because it only has \textit{(yes or no)}. Our system generates proper utterance through the following two steps for zero-shot generation.

(1) \textbf{Sim-Delexicalization}: The given unseen values are replaced with similar values among the lists of value corresponding to the same slot. In the first example of Table~\ref{Tab:zero-shot}, "Blue Man" is replaced by "Green Man" and "2.1 out of 5" is replaced by "3 out of 5". In the second example we observe that expensive is replaced by high. Also, taking into account the grammatical aspect, if an unseen value containing "the" in the \textit{\textless{}name\textgreater{}} and \textit{\textless{}near\textgreater{}} slots are given, the value list containing "the" is limited as a candidate (and vice versa). There can be several ways to find similar tokens, but we use BERTscore to select the value with the highest score.

(2) \textbf{Relexicalization}: Replaced values are changed back to unseen values in generated utterances. "Green Man" is deciphered as "Blue Man" and other values proceed as well.

The generated utterances are of appropriate quality from a human perspective. In the previous study, unlike delexicalization of unseen values to one placeholder, we have the difference of converting to similar values. Changing to one placeholder in the test also has the same risk as in training above, so we used the existing list of values to change it to an appropriate value each time. In other words, our system can generate utterances that are suitably customized for a given $(s_i, v_i)$. 
Rather than using only BERTscore, it may be helpful to find similar words using word embedding techniques such as Glove~\citep{glove} and FastText~\citep{bojanowski_etal_2017_enriching} but this will be left for further study.

\subsection{Generation from Small MRs}
\label{sec:small-MRs}
\begin{table*}[]
\centering
\resizebox{1.5\columnwidth}{!}{
\begin{tabular}{|c|c|}
\hline
{\color[HTML]{000000} Slot}                & {\color[HTML]{000000} \textless{}priceRange\textgreater{}}                                                       \\ \hline
{\color[HTML]{000000} Value}               & {\color[HTML]{000000} more   than £30'}                                                                          \\ \hline
{\color[HTML]{000000} Generated Utterance} & {\color[HTML]{000000} The   price range is more than £30. The customer rating is {\color[HTML]{FF0000} high}.}                          \\ \hline\hline

{\color[HTML]{000000} Slots}               & {\color[HTML]{000000} \textless{}name\textgreater{},   \textless{}near\textgreater{}}                            \\ \hline
{\color[HTML]{000000} Values}              & {\color[HTML]{000000} Wildwood,   The Bakers}                                                                    \\ \hline
{\color[HTML]{000000} Generated Utterance} & {\color[HTML]{000000} Wildwood   is a restaurant located near The Bakers.}                                       \\ \hline
\end{tabular}
}
\caption{Example of utterance generated when less conditions are given to our system as one pair or two $(s_i, v_i)$ pairs. The red words are the values added by the judgment of the model. The added values are not essential for composing sentences, but it makes utterance more natural}
\label{Tab:one-or-two MR}
\end{table*}
The E2E training dataset and test dataset have an average number of 5.52 and 6.91 slots, respectively, and at least 3 slots are given. In other words, the model is not trained when only one or two slots are given and is not considered in the evaluation of the test dataset. However, in the real world, the ability to generate utterances for one or two $(s_i, v_i)$ pairs is necessary, because the conditions and conditions on the generation may be insufficient. In general, methods that need to determine the structure of a sentence in advance are difficult to cope with small $(s_i, v_i)$ pairs because each value and the entire utterance do not correspond. However, our approach is end-to-end training that does not take into account the structure of the sentence, so Table~\ref{Tab:one-or-two MR} shows the possibilities for generating small pairs.
\begin{table}[t]
\centering
\resizebox{1.0\columnwidth}{!}{
\begin{tabular}{|c|c|c|c|}
\hline
\textbf{category}                        & \textbf{subject}                          & \textbf{property}                   & \textbf{object}                            \\ \hline
\multirow{2}{*}{Airport}        & Aarhus                           & leaderName                 & Jacob\_Bundsgaard                 \\ \cline{2-4} 
                                & Aarhus\_Airport                  & cityServed                 & Aarhus                            \\ \hline\hline
\multicolumn{1}{|l|}{Reference} & \multicolumn{3}{c|}{\begin{tabular}[c]{@{}c@{}}Aarhus airport serves the city of Aarhus \\ who's leader is Jacob Bundsgaard.\end{tabular}} \\ \hline
\end{tabular}
}
\caption{Example of the WebNLG dataset. In one sample, the category is fixed as Airport, and multiple values corresponding to (subject, property, object) can be given.}
\label{Tab:WebNLG_example}
\end{table}

\subsection{Experiments on a different dataset}
\begin{table}[t]
\centering
\resizebox{1.0\columnwidth}{!}{
\begin{tabular}{|c|cccc|} 
\hline
system         & BLEU & ROUGE\_L & CIDEr  & BERTscore  \\ \hline\hline
ours           & 0.2881 & 0.4859   & 2.6784 & \textbf{0.920} \\ \hline\hline
Baseline       & 0.214  & 0.3585   & 1.6754 & 0.854 \\ \hline
Melbourne\_nmt & \textbf{0.2972} & \textbf{0.516} & \textbf{3.0245} & 0.884 \\ \hline
UPF\_pipeline  & 0.2641 & 0.5018   & 2.7679 & 0.883  \\ \hline
\end{tabular}
}
\caption{Comparison of systems evaluated with the WebNLG dataset. It was evaluated using the same library as the E2E dataset.}
\label{Tab:WebNLG experiment}
\end{table}

Our paper focuses on the E2E dataset, but for the possibility of scaling, we do a simple experiment in the WebNLG challenge task~\citep{colin_etal_2016_webnlg} similar to the E2E dataset with the same approach. The WebNLG dataset is collected from DBpedia, and the train, development, and test datasets are 6940, 872, and 1862, respectively, and examples are shown in Table~\ref{Tab:WebNLG_example}. It is significantly smaller than the E2E dataset, and MRs consisting of values corresponding to four (category, subject, property, object) slots are given as a condition. In other words, unlike E2E, the WebNLG dataset has 4 fixed slots, but multiple values can be given.

Table~\ref{Tab:WebNLG experiment} shows the experimental results, and the three comparison systems that participated in the challenge~\citep{webnlg} are as follows: (1) The baseline is a neural system trained with OpenNMT~\footnote{https://opennmt.net/}. (2) Melbourne shows the best score for all automatic evaluations in the challenge with an end-to-end LSTM with attention model. Performance is improved by preprocessing entity tagging by collecting information from DBPedia. (3) UPF-FORGe~\citep{forge} is a grammer-based NLG system and has the highest score in human evaluation through rule-based graph-transducers for syntacticization. 

The systems are evaluated in the same way as the E2E dataset and our system is better than the baseline but slightly worse than the best systems in many metrics. However, our system shows meaningful results in BERTscore and is a simple method that utilizes a pre-trained model without any rule definition, preprocessing, or other datasets. If we preprocess the data like the comparison systems above, our model can expect better performance.

\subsection{Analysis}
Section~\ref{sec:zero-shot} and ~\ref{sec:small-MRs} experimentally demonstrates that our approach is capable of zero-shot generation, a situation not found in the training. Our system can cope with small $(s_i, v_i)$ pairs without considering the structure of sentences, such as the two-stage framework, sentence planning and surface realization. The system is trained to generate appropriate utterances for $(s_i, v_i)$ without the delexicalization. Therefore, there is no need to take the ambiguous risk of replacing unseen values with a single delexicalization of placeholders. Instead we introduce sim-delexicalization, which allows the system to reflect unseen values.

The two-stage framework needs to reveal the structure of the sentence, so it is difficult to solve as the number of $(s_i, v_i)$ pairs increases. However, our approach is easily extensible for more pairs. In the WebNLG dataset, we show that it is possible to extend multiple values as well. Since only slots are replaced with special tokens and values are used as regular tokens, our system can be trained to learn the utterance corresponding to MRs without limiting the number of pairs. 

We also conducted experiments using a larger backbone model, GPT2-large, but the change in performance is small. However, if the model is trained on various and massive data with a complicated sentence structure, it is expected that the experiment results using GPT2-large as a backbone would be good.

\section{Conclusion and Future Work}
This paper presents a simple one-stage approach to generating natural utterances from flat MRs. Our system uses a pre-trained model language model to improve the performance of the system and shows that it is better than the system that won the challenge in human evaluation. Even if our model is trained with only a small amount of sampling data due to the leveraging effect, it is comparable to the previous models. Our approach is simple, effective, and easy to extend to multiple MRs (i.e. WebNLG).

Our model also enables proper zero-shot generation without additional data and preprocessing due to the effects of sim-delexicalization and pretrained LM. In the future, we will use pre-trained models to build more robust systems for various datasets. Since there is no constraint to generate values in the output, it is worth considering a method that combines with reinforcement learning or a template-like approach for stability.

\bibliographystyle{acl_natbib}
\bibliography{anthology,ranlp2021}

\begin{thebibliography}{31}
\expandafter\ifx\csname natexlab\endcsname\relax\def\natexlab#1{#1}\fi

\bibitem[{Balakrishnan et~al.(2019)Balakrishnan, Rao, Upasani, White, and
  Subba}]{balakrishnan_etal_2019_constrained}
Anusha Balakrishnan, Jinfeng Rao, Kartikeya Upasani, Michael White, and Rajen
  Subba. 2019.
\newblock \href {https://doi.org/10.18653/v1/P19-1080} {Constrained decoding
  for neural {NLG} from compositional representations in task-oriented
  dialogue}.
\newblock In \emph{Proceedings of the 57th Annual Meeting of the Association
  for Computational Linguistics}, pages 831--844, Florence, Italy. Association
  for Computational Linguistics.

\bibitem[{Bojanowski et~al.(2017)Bojanowski, Grave, Joulin, and
  Mikolov}]{bojanowski_etal_2017_enriching}
Piotr Bojanowski, Edouard Grave, Armand Joulin, and Tomas Mikolov. 2017.
\newblock \href {https://doi.org/10.1162/tacl_a_00051} {Enriching word vectors
  with subword information}.
\newblock \emph{Transactions of the Association for Computational Linguistics},
  5:135--146.

\bibitem[{Colin et~al.(2016)Colin, Gardent, M{'}rabet, Narayan, and
  Perez-Beltrachini}]{colin_etal_2016_webnlg}
Emilie Colin, Claire Gardent, Yassine M{'}rabet, Shashi Narayan, and Laura
  Perez-Beltrachini. 2016.
\newblock \href {https://doi.org/10.18653/v1/W16-6626} {The {W}eb{NLG}
  challenge: Generating text from {DBP}edia data}.
\newblock In \emph{Proceedings of the 9th International Natural Language
  Generation conference}, pages 163--167, Edinburgh, UK. Association for
  Computational Linguistics.

\bibitem[{Denkowski and Lavie(2014)}]{denkowski2014meteor}
Michael Denkowski and Alon Lavie. 2014.
\newblock Meteor universal: Language specific translation evaluation for any
  target language.
\newblock In \emph{Proceedings of the ninth workshop on statistical machine
  translation}, pages 376--380.

\bibitem[{Devlin et~al.(2019)Devlin, Chang, Lee, and Toutanova}]{bert}
Jacob Devlin, Ming-Wei Chang, Kenton Lee, and Kristina Toutanova. 2019.
\newblock \href {https://doi.org/10.18653/v1/N19-1423} {{BERT}: Pre-training of
  deep bidirectional transformers for language understanding}.
\newblock In \emph{Proceedings of the 2019 Conference of the North {A}merican
  Chapter of the Association for Computational Linguistics: Human Language
  Technologies, Volume 1 (Long and Short Papers)}, pages 4171--4186,
  Minneapolis, Minnesota. Association for Computational Linguistics.

\bibitem[{Du{\v{s}}ek and
  Jur{\v{c}}{\'\i}{\v{c}}ek(2016)}]{dusek_jurcicek_2016_sequence}
Ond{\v{r}}ej Du{\v{s}}ek and Filip Jur{\v{c}}{\'\i}{\v{c}}ek. 2016.
\newblock \href {https://doi.org/10.18653/v1/P16-2008} {Sequence-to-sequence
  generation for spoken dialogue via deep syntax trees and strings}.
\newblock In \emph{Proceedings of the 54th Annual Meeting of the Association
  for Computational Linguistics (Volume 2: Short Papers)}, pages 45--51,
  Berlin, Germany. Association for Computational Linguistics.

\bibitem[{Du{\v{s}}ek et~al.(2020)Du{\v{s}}ek, Novikova, and
  Rieser}]{duvsek2020evaluating}
Ond{\v{r}}ej Du{\v{s}}ek, Jekaterina Novikova, and Verena Rieser. 2020.
\newblock Evaluating the state-of-the-art of end-to-end natural language
  generation: The e2e nlg challenge.
\newblock \emph{Computer Speech \& Language}, 59:123--156.

\bibitem[{Du{\v{s}}ek et~al.(2019)Du{\v{s}}ek, Novikova, and
  Rieser}]{dusek2019e2e}
Ond\v{r}ej Du{\v{s}}ek, Jekaterina Novikova, and Verena Rieser. 2019.
\newblock \href {https://arxiv.org/abs/1901.11528} {Evaluating the
  state-of-the-art of end-to-end natural language generation: {The} {E2E} {NLG}
  {Challenge}}.
\newblock \emph{arXiv preprint arXiv:1901.11528}.

\bibitem[{Elder et~al.(2019)Elder, Foster, Barry, and
  O{'}Connor}]{elder_etal_2019_designing}
Henry Elder, Jennifer Foster, James Barry, and Alexander O{'}Connor. 2019.
\newblock \href {https://doi.org/10.18653/v1/W19-2308} {Designing a symbolic
  intermediate representation for neural surface realization}.
\newblock In \emph{Proceedings of the Workshop on Methods for Optimizing and
  Evaluating Neural Language Generation}, pages 65--73, Minneapolis, Minnesota.
  Association for Computational Linguistics.

\bibitem[{Gardent et~al.(2017)Gardent, Shimorina, Narayan, and
  Perez-Beltrachini}]{webnlg}
Claire Gardent, Anastasia Shimorina, Shashi Narayan, and Laura
  Perez-Beltrachini. 2017.
\newblock \href {https://doi.org/10.18653/v1/W17-3518} {The {W}eb{NLG}
  challenge: Generating text from {RDF} data}.
\newblock In \emph{Proceedings of the 10th International Conference on Natural
  Language Generation}, pages 124--133, Santiago de Compostela, Spain.
  Association for Computational Linguistics.

\bibitem[{Gehrmann et~al.(2018)Gehrmann, Dai, Elder, and
  Rush}]{gehrmann_etal_2018_end}
Sebastian Gehrmann, Falcon Dai, Henry Elder, and Alexander Rush. 2018.
\newblock \href {https://doi.org/10.18653/v1/W18-6505} {End-to-end content and
  plan selection for data-to-text generation}.
\newblock In \emph{Proceedings of the 11th International Conference on Natural
  Language Generation}, pages 46--56, Tilburg University, The Netherlands.
  Association for Computational Linguistics.

\bibitem[{Juraska et~al.(2018)Juraska, Karagiannis, Bowden, and
  Walker}]{juraska_etal_2018_deep}
Juraj Juraska, Panagiotis Karagiannis, Kevin Bowden, and Marilyn Walker. 2018.
\newblock \href {https://doi.org/10.18653/v1/N18-1014} {A deep ensemble model
  with slot alignment for sequence-to-sequence natural language generation}.
\newblock In \emph{Proceedings of the 2018 Conference of the North {A}merican
  Chapter of the Association for Computational Linguistics: Human Language
  Technologies, Volume 1 (Long Papers)}, pages 152--162, New Orleans,
  Louisiana. Association for Computational Linguistics.

\bibitem[{Konstas and Lapata(2013)}]{konstas2013global}
Ioannis Konstas and Mirella Lapata. 2013.
\newblock A global model for concept-to-text generation.
\newblock \emph{Journal of Artificial Intelligence Research}, 48:305--346.

\bibitem[{Lample et~al.(2019)Lample, Subramanian, Smith, Denoyer, Ranzato, and
  Boureau}]{lample2018multipleattribute}
Guillaume Lample, Sandeep Subramanian, Eric Smith, Ludovic Denoyer,
  Marc'Aurelio Ranzato, and Y-Lan Boureau. 2019.
\newblock \href {https://openreview.net/forum?id=H1g2NhC5KQ}
  {Multiple-attribute text rewriting}.
\newblock In \emph{International Conference on Learning Representations}.

\bibitem[{Lee(2020)}]{lee_2020_stable}
Joosung Lee. 2020.
\newblock \href {https://www.aclweb.org/anthology/2020.inlg-1.25} {Stable style
  transformer: Delete and generate approach with encoder-decoder for text style
  transfer}.
\newblock In \emph{Proceedings of the 13th International Conference on Natural
  Language Generation}, pages 195--204, Dublin, Ireland. Association for
  Computational Linguistics.

\bibitem[{Lin(2004)}]{rouge}
Chin-Yew Lin. 2004.
\newblock \href {https://www.aclweb.org/anthology/W04-1013} {{ROUGE}: A package
  for automatic evaluation of summaries}.
\newblock In \emph{Text Summarization Branches Out}, pages 74--81, Barcelona,
  Spain. Association for Computational Linguistics.

\bibitem[{Lin and Och(2004)}]{lin2004automatic}
Chin-Yew Lin and Franz~Josef Och. 2004.
\newblock Automatic evaluation of machine translation quality using longest
  common subsequence and skip-bigram statistics.
\newblock In \emph{Proceedings of the 42nd Annual Meeting on Association for
  Computational Linguistics}, page 605. Association for Computational
  Linguistics.

\bibitem[{Liu et~al.(2020)Liu, Ott, Goyal, Du, Joshi, Chen, Levy, Lewis,
  Zettlemoyer, and Stoyanov}]{liu2020roberta}
Yinhan Liu, Myle Ott, Naman Goyal, Jingfei Du, Mandar Joshi, Danqi Chen, Omer
  Levy, Mike Lewis, Luke Zettlemoyer, and Veselin Stoyanov. 2020.
\newblock \href {https://openreview.net/forum?id=SyxS0T4tvS} {Ro{\{}bert{\}}a:
  A robustly optimized {\{}bert{\}} pretraining approach}.

\bibitem[{Logeswaran et~al.(2018)Logeswaran, Lee, and
  Bengio}]{logeswaran2018content}
Lajanugen Logeswaran, Honglak Lee, and Samy Bengio. 2018.
\newblock Content preserving text generation with attribute controls.
\newblock In \emph{Advances in Neural Information Processing Systems}, pages
  5103--5113.

\bibitem[{Loshchilov and Hutter(2019)}]{loshchilov2018decoupled}
Ilya Loshchilov and Frank Hutter. 2019.
\newblock \href {https://openreview.net/forum?id=Bkg6RiCqY7} {Decoupled weight
  decay regularization}.
\newblock In \emph{International Conference on Learning Representations}.

\bibitem[{Mille et~al.(2017)Mille, Carlini, Burga, and Wanner}]{forge}
Simon Mille, Roberto Carlini, Alicia Burga, and Leo Wanner. 2017.
\newblock \href {https://doi.org/10.18653/v1/S17-2158} {{FORG}e at
  {S}em{E}val-2017 task 9: Deep sentence generation based on a sequence of
  graph transducers}.
\newblock In \emph{Proceedings of the 11th International Workshop on Semantic
  Evaluation ({S}em{E}val-2017)}, pages 920--923, Vancouver, Canada.
  Association for Computational Linguistics.

\bibitem[{Nayak et~al.(2017)Nayak, Hakkani-T{\"u}r, Walker, and
  Heck}]{nayak2017plan}
Neha Nayak, Dilek Hakkani-T{\"u}r, Marilyn~A Walker, and Larry~P Heck. 2017.
\newblock To plan or not to plan? discourse planning in slot-value informed
  sequence to sequence models for language generation.
\newblock In \emph{INTERSPEECH}, pages 3339--3343.

\bibitem[{Papineni et~al.(2002)Papineni, Roukos, Ward, and
  Zhu}]{papineni2002bleu}
Kishore Papineni, Salim Roukos, Todd Ward, and Wei-Jing Zhu. 2002.
\newblock Bleu: a method for automatic evaluation of machine translation.
\newblock In \emph{Proceedings of the 40th annual meeting on association for
  computational linguistics}, pages 311--318. Association for Computational
  Linguistics.

\bibitem[{Pennington et~al.(2014)Pennington, Socher, and Manning}]{glove}
Jeffrey Pennington, Richard Socher, and Christopher Manning. 2014.
\newblock \href {https://doi.org/10.3115/v1/D14-1162} {{G}lo{V}e: Global
  vectors for word representation}.
\newblock In \emph{Proceedings of the 2014 Conference on Empirical Methods in
  Natural Language Processing ({EMNLP})}, pages 1532--1543, Doha, Qatar.
  Association for Computational Linguistics.

\bibitem[{Puzikov and Gurevych(2018)}]{puzikov_gurevych_2018_e2e}
Yevgeniy Puzikov and Iryna Gurevych. 2018.
\newblock \href {https://doi.org/10.18653/v1/W18-6557} {{E}2{E} {NLG}
  challenge: Neural models vs. templates}.
\newblock In \emph{Proceedings of the 11th International Conference on Natural
  Language Generation}, pages 463--471, Tilburg University, The Netherlands.
  Association for Computational Linguistics.

\bibitem[{Radford et~al.(2019)Radford, Wu, Child, Luan, Amodei, and
  Sutskever}]{radford2019language}
Alec Radford, Jeffrey Wu, Rewon Child, David Luan, Dario Amodei, and Ilya
  Sutskever. 2019.
\newblock Language models are unsupervised multitask learners.
\newblock \emph{OpenAI Blog}, 1(8):9.

\bibitem[{Smiley et~al.(2018)Smiley, Davoodi, Song, and
  Schilder}]{smiley_etal_2018_e2e}
Charese Smiley, Elnaz Davoodi, Dezhao Song, and Frank Schilder. 2018.
\newblock \href {https://doi.org/10.18653/v1/W18-6558} {The {E}2{E} {NLG}
  challenge: A tale of two systems}.
\newblock In \emph{Proceedings of the 11th International Conference on Natural
  Language Generation}, pages 472--477, Tilburg University, The Netherlands.
  Association for Computational Linguistics.

\bibitem[{Vaswani et~al.(2017)Vaswani, Shazeer, Parmar, Uszkoreit, Jones,
  Gomez, Kaiser, and Polosukhin}]{vaswani2017attention}
Ashish Vaswani, Noam Shazeer, Niki Parmar, Jakob Uszkoreit, Llion Jones,
  Aidan~N Gomez, {\L}ukasz Kaiser, and Illia Polosukhin. 2017.
\newblock Attention is all you need.
\newblock In \emph{Advances in neural information processing systems}, pages
  5998--6008.

\bibitem[{Vedantam et~al.(2015)Vedantam, Lawrence~Zitnick, and
  Parikh}]{vedantam2015cider}
Ramakrishna Vedantam, C~Lawrence~Zitnick, and Devi Parikh. 2015.
\newblock Cider: Consensus-based image description evaluation.
\newblock In \emph{Proceedings of the IEEE conference on computer vision and
  pattern recognition}, pages 4566--4575.

\bibitem[{Wiseman et~al.(2018)Wiseman, Shieber, and
  Rush}]{wiseman_etal_2018_learning}
Sam Wiseman, Stuart Shieber, and Alexander Rush. 2018.
\newblock \href {https://doi.org/10.18653/v1/D18-1356} {Learning neural
  templates for text generation}.
\newblock In \emph{Proceedings of the 2018 Conference on Empirical Methods in
  Natural Language Processing}, pages 3174--3187, Brussels, Belgium.
  Association for Computational Linguistics.

\bibitem[{Zhang* et~al.(2020)Zhang*, Kishore*, Wu*, Weinberger, and
  Artzi}]{Zhang2020BERTScore}
Tianyi Zhang*, Varsha Kishore*, Felix Wu*, Kilian~Q. Weinberger, and Yoav
  Artzi. 2020.
\newblock \href {https://openreview.net/forum?id=SkeHuCVFDr} {Bertscore:
  Evaluating text generation with bert}.
\newblock In \emph{International Conference on Learning Representations}.

\end{thebibliography}


\end{document}